# Multi-Timescale Conductance Spiking Networks: A Sparse, Gradient-Trainable Framework with Rich Firing Dynamics for Enhanced Temporal Processing

Alex Fulleda-Garcia*, Saray Soldado-Magraner†‡, Josep Maria Margarit-Taulé*‡

**Instituto de Microelectrónica de Barcelona (IMB-CNM),*
*Consejo Superior de Investigaciones Científicas (CSIC)*
Cerdanyola del Vallès, Spain
alex.fulleda@csic.es, josepmaria.margarit@csic.es

†*Department of Neurobiology,*
*University of California Los Angeles (UCLA)*
Los Angeles, USA
ssaray@ucla.edu

‡*Joint Senior authors*

***Abstract*—Spiking neural networks (SNNs) promise low-power event-driven computation for temporally rich tasks, but commonly used neuron models often trade off gradient-based trainability, dynamical richness, and high activity sparsity. These limitations are acute in regression, where approximation error, noise and spike discretization can severely degrade continuous-valued outputs. Indeed, many state-of-the-art (SOTA) SNNs rely on simple phenomenological dynamics trained with surrogate gradients and offer limited control over spiking diversity and sparsity. To overcome such limitations, we introduce multi-timescale conductance spiking networks, a gradient-trainable framework in which neural dynamics emerge from shaping the current-voltage (I–V) curve by tuning fast, slow and ultra-slow conductances. This parametrization allows systematic control over excitability, can be implemented efficiently in analog circuits, and yields rich firing regimes including tonic, phasic and bursting responses within a single model. We derive a discrete-time formulation of these differentiable dynamics, enabling direct backpropagation through time without surrogate-gradient approximations. To probe both trainability and accuracy, we evaluate feedforward networks of these neurons at the predictability limit of Mackey–Glass time-series regression and compare them to baseline LIF and SOTA AdLIF networks. Our model outperforms LIF and AdLIF networks, while exhibiting substantially sparser activity from both communication and computational perspectives. These results highlight multi-timescale conductance spiking neurons as a promising building block for energy-aware temporal processing and neuromorphic implementation.**

***Index Terms*—Spiking Neural Networks (SNNs), Conductance-Based Neuron Models, Multi-Timescale Dynamics, Temporal Regression, Neuromorphic Computing**

## I. Introduction (and Related Work)

Spiking neural networks (SNNs) have emerged as a compelling substrate for energy-efficient computation. By leveraging stateful neuron dynamics and event-driven communication, SNNs offer intrinsic memory and sparse signaling capabilities that map naturally onto neuromorphic hardware. While SNNs hold promise to outperform conventional Artificial Neural Networks (ANNs) on time-dependent tasks by mimicking temporal processing in the brain, the gap between biological plausibility and machine learning trainability remains a significant hurdle.

Moreover, realizing this potential requires moving beyond simple accuracy metrics on static datasets. There is a growing consensus within the community about the importance of fair and standardized benchmarks that prioritize task capabilities relevant to neuromorphic approaches—specifically temporal processing—and efficiency metrics such as sparsity and energy consumption [1], [2]. While most existing neuromorphic benchmarks and SNN studies focus on event-based classification tasks, continuous-time regression problems with long-range temporal dependencies—which are less tolerant to approximation error and more sensitive to noise and spike quantization—remain comparatively underexplored. However, continual regression is crucial for many practical applications,

This work was partially supported by a Ramón y Cajal Fellowship (grant RYC2022-036701-I) and a Consolidación Investigadora grants (grant CNS2023-143734) to JMMT, funded by MCIN/AEI/10.13039/501100011033 and by the European Union "NextGenerationEU"/PRTR, as well as Swiss National Science Foundation grants P2ZHP3-187943/P500PB-203133, and a Neurotech 2019 grant to SSM. This project has also received funding from the European Union's Horizon Europe research and innovation program under grant agreement Nr 101135241 (BIOSENSEI). We gratefully acknowledge the computing resources provided by Artemisa at the Instituto de Física Corpuscular (IFIC, CSIC–UV) and by the FinisTerrae III supercomputer at the Galician Supercomputing Center (CESGA), the first funded by the Generalitat Valenciana and both by the European Union ERDF the European Union "NextGenerationEU"/PRTR. We finally thank the Capocaccia Workshop for Neuromorphic Intelligence, where this project was conceived.

Corresponding author: josepmaria.margarit@csic.es

including industrial monitoring [3] and fault prediction [4], physiological tracking [5], and real-time closed-loop control [6], where long-term accuracy and robustness are critical. Demonstrating true advantage for these applications requires a holistic evaluation that balances algorithmic complexity with hardware constraints.

Currently, the vast majority of machine-learning-oriented SNNs rely on highly simplified neuron dynamics—predominantly the Leaky Integrate-and-Fire (LIF) model and its variants [7]–[9]. These phenomenological models sacrifice biophysical realism in favor of computational efficiency, by replacing detailed conductance dynamics with abstract variables that approximate spiking behavior [10], [11] as depicted in Fig. 1(a). To overcome the non-differentiability of the hard spike threshold, phenomenological models are typically trained using surrogate gradient techniques [12], [13] approximating the backward pass derivative with a smoothed function (e.g., a sigmoid or arctan).

Although this combination has enabled deep SNN training and competitive performance on standard benchmarks, it comes with several limitations that become especially pronounced for continuous-valued temporal regression. First, LIF-based neurons strip away the rich intrinsic mechanisms that endow biological neurons with high-dimensional temporal processing capabilities, offering limited control over intrinsic dynamics through a small number of tunable time constants and a restricted repertoire of firing regimes. Second, the use of surrogate gradients introduces a mismatch between the forward dynamics and the backward pass—limiting the faithful learning of complex temporal dynamics when optimizing sensitive regression losses—and often requires careful tuning of surrogate shapes and training hyperparameters. Finally, sparsity in these models tends to be managed indirectly via threshold or loss regularization, rather than emerging from explicit, interpretable mechanisms in the single-neuron dynamics. Consequently, many SNNs struggle to capture complex sequential dependencies without recurrent inter-neuron architectural overhead.

Recently, the field has recognized the importance of restoring dynamical realism to SNNs. Multiple studies have demonstrated that incorporating spike-frequency adaptation expands the computational power of spiking networks, particularly for temporal dependencies [14]–[16]. Modern adaptive-LIF variants extend this principle, showing strong performance gains by introducing a secondary sub-threshold variable that regulates the dynamics of the membrane potential [17], [18]. These results support the idea that intrinsic state variables evolving over multiple timescales can play a critical role in temporal processing.

However, even these adaptive variants capture only a narrow subset of the diversity seen in biological systems. The rich firing modes present in real neurons arise not merely from a secondary slow state variable, but from the complex interplay of ion channel conductances operating at multiple timescales [19], [20]. Importantly, the conductances underlying these modes are not static; they attune to input statistics, dynamically reshaping the neuron's effective I–V curve [21], [22]. Such a feature endows the neuron with history-dependent integration properties that promote temporally sparse patterns of activity. This body of work suggests that intrinsic conductance modulation is itself a computational mechanism that biological circuits exploit.

Recent neuromorphic circuit research has developed compact realizations of these complex neural dynamics. Ribar and Sepulchre introduced a reduced conductance-based framework in which the behavior of a neuromorphic circuit is controlled by shaping its current–voltage (I–V) curve via a parallel interconnection of positive and negative conductance elements operating at different time scales [23] as the (voltage-controlled) current sources $I_x$ of Fig. 1(b). In this framework, the excitability and firing regime of a neuron-like circuit are determined primarily by the slopes and intersections of aggregate I–V curves composed of fast, slow, and ultra-slow conductances. Such multi-timescale, geometrically interpretable parameterizations can capture rich spiking behaviors and provide smooth internal representations that mitigate the effects of spike discretization, while remaining amenable to analysis and efficient analog circuit design [24]: localized conductance elements can be implemented with compact transconductance blocks (e.g., subthreshold MOS) whose I–V characteristics need not be exact tanh functions, but can be any approximately monotone nonlinearity of a filtered voltage [25], [26].

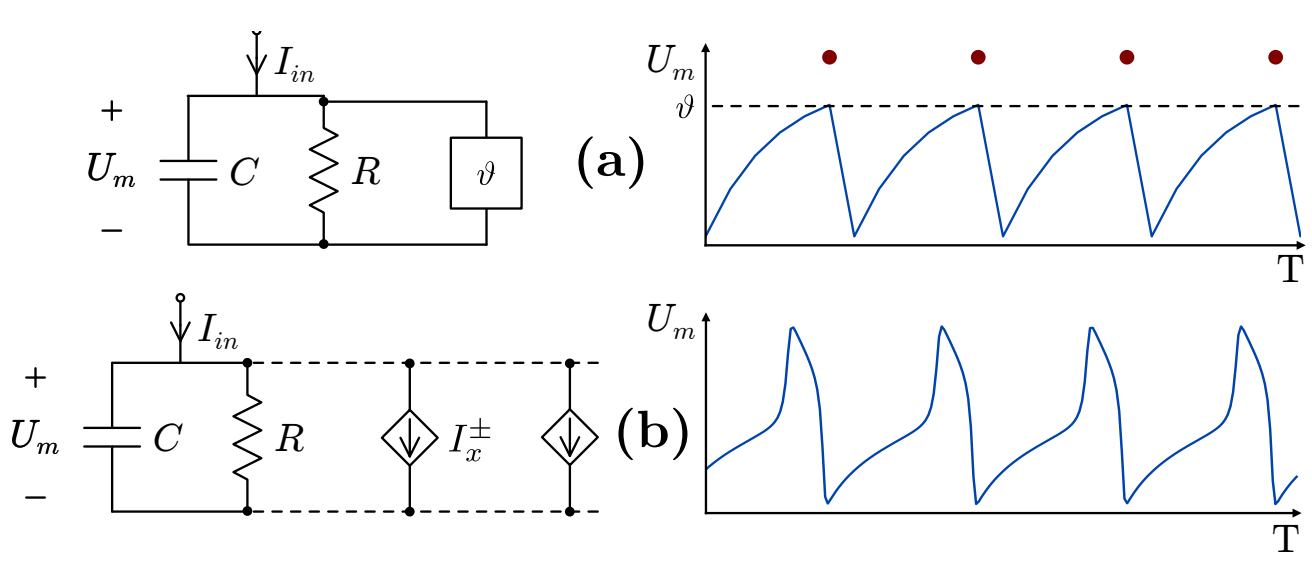

Fig. 1. Equivalent electrical circuit of a phenomenological integrate-and-fire model (a, left), and threshold-crossing discontinuities exhibited when generating spikes and resetting its membrane voltage (a, right). Equivalent electrical circuit of a reduced multi-timescale conductance-based model (b, left), and its characteristic continuous spiking response (b, right).

Here, we explore the computational properties of such rich spiking dynamics in machine-learning settings. Building on the compact conductance-based parametrization of fast, slow, and ultra-slow conductance elements from [23], we derive a discrete-time, differentiable formulation that yields a continuum of firing behaviors—including tonic spiking, phasic spiking, and a diversity of bursting behaviors. SNNs of this framework, to which we refer as Multi-Timescale Conductance (MTC), are directly trainable with gradient descent and backpropagation through time (BPTT), without relying on surrogate gradients.

We benchmark a feedforward MTC-SNN against LIF and state-of-the-art (SOTA) Adaptive LIF (AdLIF) baselines in terms of correctness and activity sparsity metrics when directly forecasting a regression problem requiring long-range temporal structure and precise continuous-valued outputs: a Mackey–Glass time series [27] at its predictability horizon. Our results show that conductance-based units achieve improved accuracy over LIF models and SOTA adaptive baselines, while operating in a considerably sparser dynamic regime across both rate and duty-cycle dimensions. This aligns with the broader neuromorphic vision that dynamic sparsity—implemented here at the level of single-neuron excitability—can be a key lever to deliver true neuromorphic advantage in low-power intelligent perception [28].

Taken together, these findings highlight the value of biologically grounded conductance mechanisms for temporal processing and identify conductance-shaped excitability as a promising path toward energy-efficient and temporally expressive spiking architectures.

## II. Methodology

### A. Problem Formulation: Time Series Forecasting

We formulate the time series forecasting task as a nonlinear regression problem. Given a historical sequence $X_t = \{x_t, x_{t-1}, \dots, x_{t-T+1}\}$ of length $T$, the objective is to learn a mapping function $\mathcal{F}$ that predicts a future value $x_{t+d}$:

$$\hat{x}_{t+d} = \mathcal{F}(X_t; \theta)$$

where $\theta$ represents the trainable parameters of the network and $d \geq 1$ denotes the prediction horizon. In this work, we specifically focus on multi-step-ahead prediction (where $d > 1$), a regime that rigorously tests the model's ability to capture long-range temporal dependencies.

### B. Integrate-and-Fire Spiking Neural Networks

To establish a baseline, we consider the standard Leaky Integrate-and-Fire (LIF) model. The LIF neuron models the membrane potential $U_m(t)$ as a generic RC circuit driven by an input current $I_{in}(t)$:

$$\tau_m \frac{dU_m(t)}{dt} = -(U_m(t) - U_{rest}) + RI_{in}(t) \tag{1}$$

where $\tau_m$ is the membrane time constant and $R$ is the membrane resistance. When $U_m(t)$ reaches a firing threshold $U_{th}$, the neuron emits a discrete spike $S(t) = \delta(t - t_{spike})$ and the potential is instantaneously reset to a value $U_{reset}$ typically equal to $U_{rest}$. While the LIF model is computationally efficient, it lacks the rich dynamical repertoire of biological neurons. To boost performance in temporal tasks, recent approaches often incorporate mechanisms of spike frequency adaptation. This is the case of the SOTA AdLIF model [18], which augments the state space with an adaptation variable $W(t)$, creating a coupled system:

$$\tau_m \frac{dU_m(t)}{dt} = -(U_m(t) - U_{rest}) + RI_{in}(t) - RW(t) \tag{2}$$

$$\tau_w \frac{dW(t)}{dt} = a(U_m(t) - U_{rest}) - W(t) + b\sum_k \delta(t - t_k) \tag{3}$$

Here, the variable $W(t)$ implements a negative subthreshold feedback loop that is believed to effectively extend the neuron's temporal integration window and dynamic range.

### C. Multi-Timescale Conductance Spiking Neural Networks

To overcome the limitations of surrogate gradients and fixed dynamics, we develop the conductance-based neuron model proposed by Ribar and Sepulchre [23] into a discrete-time implementation of a differentiable Multi-Timescale Conductance (MTC) SNN. Unlike the hybrid continuous-discrete nature of LIF models, our MTC-SNN framework produces spikes through fully derivable nonlinear dynamics, driven by conductance elements operating at distinct timescales to emulate biological ionic currents.

The base layer of the model is an RC circuit like in Eq. 1. The dynamics of the voltage-gated conductance elements $I_x$ are modeled as:

$$\tau_x \frac{dU_x(t)}{dt} = -U_x(t) + U_m(t) \tag{4}$$

$$I_x^{\pm}(t) = \pm\alpha_x^{\pm} \tanh(U_x(t) - \delta_x^{\pm}), \tag{5}$$

where $x$ is an index indicating the timescale of the conductance, $\tau_x$ is the time constant governing the delay of the state variable $U_x$ relative to the membrane potential, $\alpha_x^{\pm}$ represents the maximal conductance (gain) of the channel, and $\delta_x^{\pm}$ determines the voltage range in which the element is active.

The total membrane potential dynamics are governed by the integration of these voltage-gated currents alongside the passive leak and the external input:

$$\tau_m \frac{dU_m(t)}{dt} = -(U_m(t) - U_{rest}) + RI_{in}(t) - R\sum_{x\in X} I_x^{\pm}(t) \tag{6}$$

For numerical simplicity, we normalize the circuit resistance $R = 1$, effectively scaling currents relative to the leak conductance.

The LIF-based systems are discretized with the method employed in their respective reference implementations [29], [18]: Explicit Exponential Euler for the LIF and Symplectic Exponential Euler for the AdLIF model. Our MTC model is discretized using the Explicit Euler-Forward:

$$U_x[t+1] = U_x[t] + \frac{dt}{\tau_x}(-U_x[t] + U_m[t]) \tag{7}$$

$$U_m[t+1] = U_m[t] + \frac{dt}{\tau_m}\left(-U_m[t] + U_{rest} + I_{in}[t] - \sum I_x^{\pm}[t]\right) \tag{8}$$

The conductance elements are organized into three distinct timescales relative to the membrane time constant, $\tau_m$: fast ($\tau_f$), slow ($\tau_s$), and ultra-slow ($\tau_{us}$). Following the circuit-theoretic construction of excitability [23], two primary elements generate the spiking limit cycle. The first one is a fast element $I_f^-$ with negative conductance. By assuming the fast dynamics are instantaneous relative to the membrane ($\tau_f \to 1$), this element creates a region of negative differential

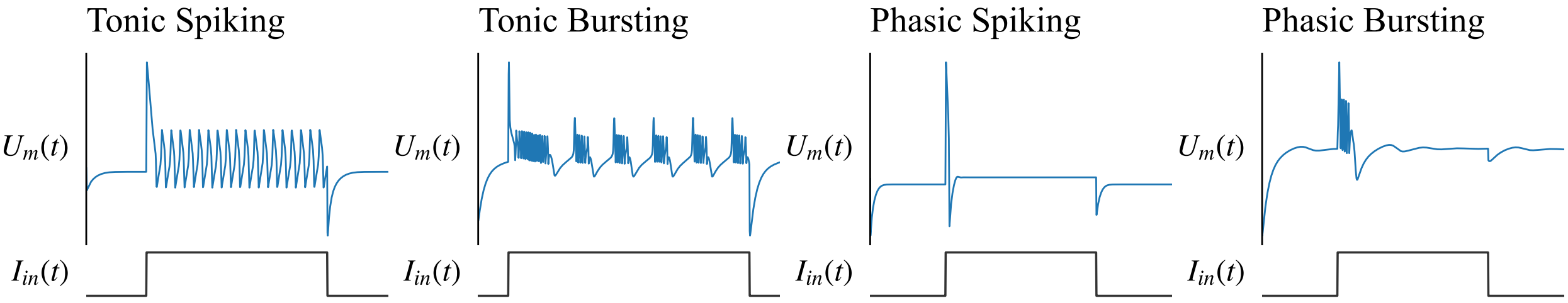

Fig. 2. Diversity of neuronal firing patterns. The figure illustrates four distinct firing regimes generated by the neuron model in response to external stimulation. In each panel, the upper trace represents the membrane potential, $U_m$, and the lower trace indicates the input current, $I_{in}(t)$. The patterns include Tonic Spiking and Tonic Bursting (sustained responses to constant input), Phasic Spiking and Phasic Bursting (transient responses to input onset)

resistance in the system's I–V curve. Such an instability injects energy into the circuit, driving the rapid depolarization (upstroke) of the action potential. Secondly, a restorative element $I_s^+$ with positive conductance and a slower time constant ($\tau_s \gg \tau_m$) provides the necessary damping force to recover the membrane potential after a spike and enforces the refractory period.

Beyond simple spiking, an ultra-slow timescale ($\tau_{us} \gg \tau_s$) enables higher-order temporal processing. By adding a slow-negative element $I_s^-$, a second negative conductance region is created, now on the slow timescale. This new slow element is balanced by a positive conductance $I_{us}^+$ in the ultra-slow timescale. Modulating the parameters of these elements allows the model to smoothly transition between distinct firing modes, such as, but not limited to, tonic spiking (constant firing) and bursting (clusters of spikes followed by silence), as can be observed in Fig. 2.

While the conductance-based framework naturally yields a continuous voltage trajectory $U_m(t)$ that exhibits spiking behavior, directly utilizing the raw membrane potential for inter-neuron communication or readout introduces significant computational challenges. In biological systems, the axonal propagation and subsequent mechanisms of neurotransmission at the synapse could be seen as a threshold-based rectifying filter, converting the analog membrane fluctuations into pulsed neurotransmitter release. Similarly, in a computational context, relying on raw voltage couples the network to sub-threshold noise and amplitude variations, potentially destabilizing the readout layer. Thus, we implement a signal conditioning layer that transforms the internal membrane state into a standardized transmission signal $s(t)$. This is achieved via a normalized Saturated ReLU activation function:

$$s(t) = \min\left(\frac{\text{ReLU}(U_m(t) - U_{th})}{U_{sat} - U_{th}}, 1\right) \tag{9}$$

where $U_{th}$ represents the firing threshold and $U_{sat}$ corresponds to the peak saturation voltage.

Such a formulation helps on two primary axes: The first is signal standardization and stability. The raw action potentials generated by the conductance model can vary in amplitude depending on input intensity and parameter tuning. The readout function normalizes these events to the unit interval $[0, 1]$. This amplitude simplifies network scalability by ensuring that downstream layers or recurrent connections receive consistent input magnitudes regardless of the varying internal dynamics of the source neurons. The second is semi-digital communication. For efficient neuromorphic operation, it is desirable to suppress sub-threshold activity while maintaining the differentiability of the spike onset. This function acts as a noise gate, forcing $s(t) = 0$ for all activity below $U_{th}$. The resulting signal is "semi-digital": it creates a sparse, high-contrast communication channel similar to binary spikes, yet retains the continuous slope information during the rising phase ($U_{th} < U_m < U_{sat}$), which is necessary for calculating exact gradients during training.

Eq. 9 works as a Synaptic Transduction Model. Physiologically, this mapping approximates the non-linear relationship between pre-synaptic voltage and neurotransmitter release. Synaptic transmission is not a linear function of voltage; it requires a minimum depolarization to trigger vesicle fusion ($U_{th}$) and saturates upon depletion of the readily releasable pool ($U_{sat}$). By modeling this transduction step, we enforce the network to communicate via "events" rather than continuous voltage leakage.

## III. Experiments and Results

This section presents the comparative performance of the models in the chaotic Mackey–Glass forecasting task. The experimental setup and procedures are introduced before continuing with the outcome of the experiments for all MTC, LIF, and SOTA AdLIF networks.

### A. Experimental Setup

*1) Dataset:* We employ the Mackey–Glass (MG) chaotic time series [27], a standard benchmark for reservoir computing and neuromorphic forecasting tasks [30], [31]. The series is generated using the standard parameters $\gamma = 0.1$, $\beta = 0.2$, $n = 10$ and a delay $\tau = 17$, which places the system in a chaotic regime.

$$\frac{dx(t)}{dt} = \beta \frac{x(t-\tau)}{1 + x(t-\tau)^n} - \gamma x(t) \tag{10}$$

The MG differential equation is numerically integrated using the Forward Euler method with a discretization step of $dt = 0.2$. This sampling rate was selected according to the slow intrinsic dynamics of our MTC neuron model. To ensure that the system trajectory has relaxed onto the chaotic attractor, an initial transient period of 1000 time units (5000 steps) is discarded.

Following transient removal, the time series is normalized using a MinMax scaler constrained to the range $[-0.5, 0.5]$. This zero-centered interval was empirically selected to optimally engage the dynamic range of the neuron models.

For this supervised learning task, the processed series is segmented into input-target pairs. Each input sample consists of a history window of length $T_x$, and the corresponding target is a window of the same duration, temporally displaced by a prediction horizon of $\Delta t = 675$ timesteps. This horizon corresponds to approximately one Lyapunov time for the system[1], representing the critical point where the system's sensitivity to initial conditions renders prediction highly non-linear and challenging [32].

*2) Comparative Baselines:* To isolate the performance benefits of the proposed conductance-based dynamics, we benchmark our model against two widely used spiking neuron formulations: the LIF and AdLIF models. The Leaky Integrate-and-Fire serves as the standard non-adaptive baseline. To ensure a robust implementation, we applied the optimized modules provided by the `snnTorch` framework [7], [29]. The AdLIF model represents the SOTA for simple adaptive spiking models, incorporating a secondary state variable that implements a negative subthreshold feedback current. The implementation was adapted directly from the reference code provided in [18], [33], ensuring that the baseline reflects established performance standards. This comparison allows us to distinguish between gains arising from simple integration (LIF) versus those derived from standard adaptation mechanisms (AdLIF).

*3) Network architecture:* All models utilize a feed-forward spiking architecture with no lateral or recurrent synaptic connections. The topology consists of four stages: a linear input projection ($1 \times N$) that maps the scalar time series into a high-dimensional space; a hidden processing layer containing $N$ independent spiking neurons; a readout layer ($N \times 1$) that linearly decodes the spike trains; and a 4th-order low-pass filter, to reconstruct the continuous target signal from the discrete output spikes and mitigate high-frequency quantization noise. It is important to note that while the network topology is feed-forward, the system possesses intrinsic memory. The temporal dependencies of the input signal are integrated via the internal dynamics of the state variables within each neuron, enabling the processing of time-series data without the need for network-level recurrence.

[1]To validate this horizon, we performed a phase space reconstruction using the Mutual Information method for delay selection ($\tau_{embed} = 12$) and False Nearest Neighbors for the embedding dimension ($m = 4$). Using the Rosenstein algorithm, we estimated the maximal Lyapunov exponent for our generated series, yielding a Lyapunov time of approximately 135 time units. Consequently, our chosen horizon of 675 timesteps ($135/0.2$) corresponds to one Lyapunov time, pushing the model to the limit of deterministic predictability.

*4) Training details and Hyperparameters:* To ensure a rigorous and reproducible evaluation, we adopted a structured multi-stage optimization protocol. The training configuration encompasses the optimization framework, gradient estimation strategies, weight initialization, and a hierarchical hyperparameter search.

*a) Optimization Framework:* All models were trained for a fixed duration of 10,000 epochs to ensure convergence. The dataset was fed on batches of 128 samples each. We utilized the Adam optimizer with default parameters ($\beta_1 = 0.9, \beta_2 = 0.999$). To stabilize training dynamics, we employed a Cosine Annealing learning rate scheduler, which decays the learning rate following a cosine curve over the course of training. The optimization objective was to minimize the Mean Squared Error (MSE) between the filtered output signal and the ground truth target.

*b) Gradient Estimation:* Given the non-differentiable nature of spike generation, different strategies were employed for backward pass validation:

- **LIF:** Trained using the default surrogate gradient mechanism provided by `snnTorch` (ArcTan derivative).
- **AdLIF:** Utilized the SLAYER surrogate gradient [13] with a smoothing factor $\alpha = 5$, consistent with the reference implementation.
- **MTC:** Unlike the baselines, our conductance-based model was trained using standard Backpropagation Through Time (BPTT) without surrogate gradients, exploiting the differentiable nature of the continuous conductance state variables.

*c) Weight Initialization:* To accommodate the specific architectural requirements of each neuron model, distinct initialization schemes were applied:

- **LIF and MTC:** The input projection layer was initialized using He Uniform initialization to maintain variance stability. The linear readout layer employed LeCun Normal initialization.
- **AdLIF:** Followed the specific protocol from the reference literature [18]. Input and output weights were initialized via a scaled uniform distribution bounded by $\pm\text{gain} \times \sqrt{1/\text{fan_in}}$.

*d) Hyperparameter Search Strategy:* We implemented a "fair comparison" search strategy divided into three phases:

1) **Neuron Dynamics Tuning:** For all models, we performed a grid search over key intrinsic parameters (decay rates (LIF); adaptation time constants and feed-forward gain (AdLIF); and conductance time constants (MTC)) to minimize validation error. For our MTC model, other parameters were manually tuned by analyzing the phase-space dynamics, I–V curves and $U_m$ traces of individual neurons under varying input currents to ensure rich temporal behavior.
2) **Architecture and Global Parameters:** A random search was conducted over the macro-parameters: input window size ($T_x$), dataset size ($N_{samples}$), hidden layer dimension ($N$), and base learning rate.

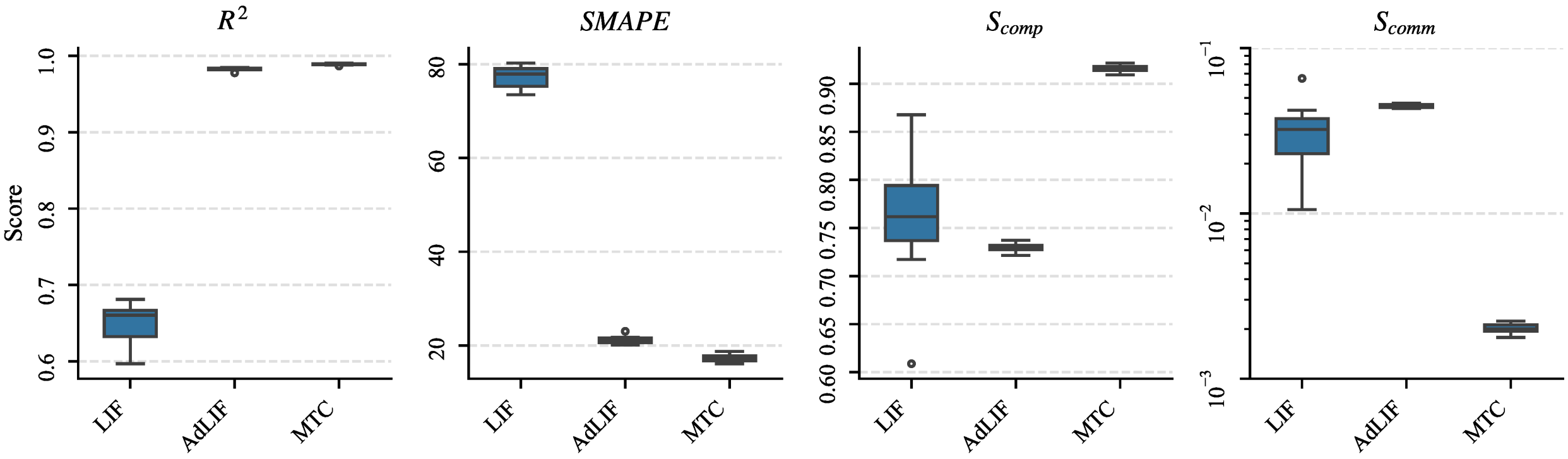

Fig. 3. Comparative evaluation of forecasting fidelity and efficiency. Boxplots summarize performance metrics over 10 independent trials. Left panels ($R^2$, *SMAPE*): The proposed MTC architecture and the adaptive baseline (AdLIF) outperform the static LIF model, achieving near-optimal forecasting fidelity. Right panels (Sparsity, Spike Probability): Efficiency metrics reveal distinct encoding strategies. While AdLIF relies on dense firing (low sparsity, high spike probability) to maintain accuracy, the MTC model achieves the highest $S_{comp}$ (>0.92) and lowest $S_{comm}$.

3) **Final Selection:** Based on the random search results, we fixed the structural parameters ($T_x = 2000$, $N_{samples} = 1500$, $N = 1000$) to be identical across all models to guarantee architectural fairness. However, the learning rate was set to the specific optimal value found for each distinct model type. Using the optimized configurations, each model was trained and evaluated over 10 random seeds. The reported results represent the mean and standard deviation across these trials.

*5) Evaluation Metrics:* We evaluate model performance along two primary axes: predictive fidelity and activity sparsity. Regarding fidelity, we employ two standard metrics for regression tasks: the coefficient of determination ($R^2$) and the Symmetric Mean Absolute Percentage Error (*SMAPE*). $R^2$ offers a standardized measure of goodness-of-fit relative to the data variance, while *SMAPE* provides a scale-independent interpretability that decouples performance from the signal amplitude. For activity sparsity, we introduce two metrics to quantify gains relevant to neuromorphic hardware.

Computational Sparsity ($S_{comp}$): This metric measures the proportion of time in which synaptic operations remain idle. For a network of $N$ neurons over $T$ timesteps, where $\mathbb{I}(\cdot)$ is the indicator function and $s_i(t)$ is the spike state of neuron $i$ at time $t$:

$$S_{comp} = \frac{1}{N \times T} \sum_{t=1}^{T} \sum_{i=1}^{N} \mathbb{I}(s_i(t) = 0) \tag{11}$$

A higher $S_{comp}$ indicates a longer duration in the 'off' state, potentially reducing dynamic power consumption related to spike processing.

Communication Sparsity ($S_{comm}$): To quantify the bandwidth usage (inter-neuron communication), we measure the average spike probability per timestep:

$$S_{comm} = \frac{\sum_{t,i} s_i(t)}{N \times T} \tag{12}$$

Here, a lower $S_{comm}$ is preferred as, in hardware terms, it implies a lower energy expenditure when transmitting the events.

### *B. Boosting Performance through Temporal Expressivity and Heterogeneous Firing*

To assess the benefits of multi-timescale conductance dynamics, we first evaluate performance on the MG task using the standard LIF model. As shown in Figure 3, the LIF model struggles to capture the complex dynamics over the distant prediction horizon (d=675), achieving a median $R^2$ of approximately 0.66 and a high *SMAPE* error of 78%. These results suggest that the single-timescale membrane integration dynamics of the standard LIF are insufficient for retaining the long-term dependencies required for this task.

We then compare a SOTA AdLIF model with its time constants distributed over the ranges $\tau_m \in [1, 5]$ and $\tau_w \in [60, 300]$, against our MTC architecture with taus also distributed over $\tau_s \in [20, 125]$ and $\tau_{us} \in [2000, 4000]$. Besides this temporal expressivity, MTC networks introduce another key physiological feature: firing diversity (50% is restrained to only tonic spiking behavior with fast-negative and slow-positive conductance ($\alpha_f^- = 2, \alpha_s^+ = 10$) while the other 50% of the population also incorporates slow-negative and ultra-slow negative conductance ($\alpha_s^- = -7, \alpha_{us}^+ = 20$)). The effects of these two added properties can be clearly observed in Fig. 4

In contrast to LIF, both adaptive models achieve excellent convergence. Our proposed MTC model achieves the highest overall accuracy (Median $R^2 \approx 0.988$, *SMAPE* $\approx 17\%$) closely followed by the AdLIF with similar predictive power (Median $R^2 \approx 0.982$, *SMAPE* $\approx 21\%$).

While the accuracy metrics are comparable, the efficiency metrics reveal a fundamental difference in how the models solve the task. The AdLIF achieves high accuracy through a dense firing strategy, exhibiting a spike rate of $S_{comm} \approx 4 \cdot 10^{-2}$. In contrast, the MTC-SNN solves the task with $S_{comm} \approx 2 \cdot 10^{-3}$– an order of magnitude lower.

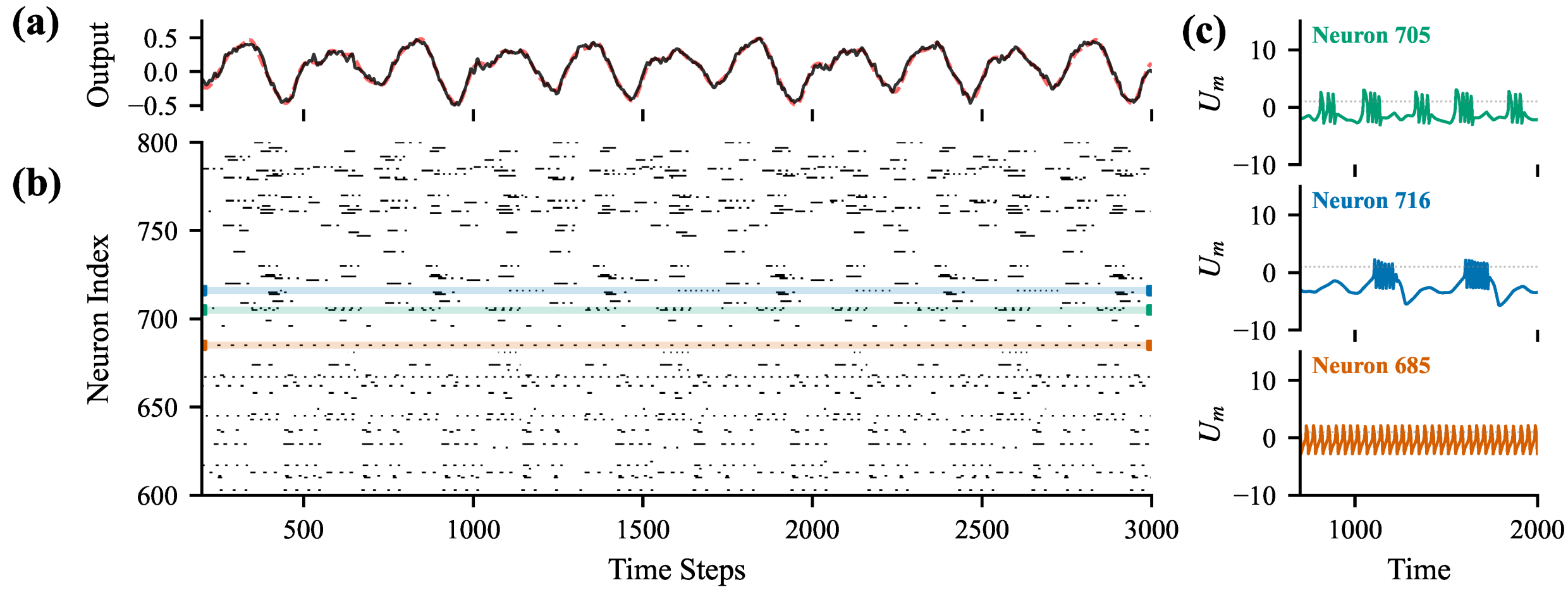

Fig. 4. Population dynamics and heterogeneous firing regimes in the MTC model. Model prediction with forecasting horizon $d = 675$ (a, black) accurately tracking the Mackey–Glass time series ground-truth target (a, red-dashed). Spike raster plot of the hidden layer activity. The network exhibits a sparse, distributed coding scheme. The shaded horizontal bands highlight three specific neurons selected to illustrate the diversity of intrinsic dynamics (b). Voltage traces ($U_m$) of selected neurons, revealing distinct firing modes emerging from the framework proposed: (c, top) high inter-burst frequency model, (c, middle) low inter-burst with high intra-burst firing frequency and (c, bottom) tonic spiking neuron.

This is further corroborated by $S_{comp}$, where the MTC network achieves the highest computational sparsity of all models ($> 0.9$). The low efficiency of the AdLIF model is in fact comparable to the LIF model in terms of $S_{comp}$ average metrics ($\approx 0.7$), but with less variance.

These results suggest that the inclusion of heterogenous firing and time constants allows the MTC model to encode information in highly compressed, temporally precise spike volleys, rather than the regular tonic spiking regimes observed in the AdLIF. Consequently, while the AdLIF offers comparable accuracy, the MTC architecture provides a far more efficient solution as to synaptic activity, reducing inter-neuron communication bandwidth requirements beyond one order of magnitude with negligible loss in predictive power.

Finally, Fig. 5 synthesizes these findings by mapping the models into a Performance-Efficiency plane. The plot shows the standard LIF baseline struggles in both axes, accuracy and sparsity, compared to all the other models. While the AdLIF is able to retain competitive performance, the MTC model achieves a higher overall accuracy score while also leading efficiency metrics, positioning our approach to the upper-right quadrant of the map.

## IV. Conclusions and Future Work

In this work, we introduced Multi-Timescale Conductance (MTC) spiking networks, a gradient-trainable conductance-based neuron framework in which rich firing dynamics emerge from shaping the current–voltage (I–V) curve via fast, slow, and ultra-slow conductance elements. By deriving a differentiable discrete-time formulation, we enable direct training with BPTT without resorting to surrogate gradients, while retaining a geometrically interpretable, biophysically motivated parametrization of excitability and firing regime. Within a single neuron model, the same parametrization can express tonic spiking, phasic and diverse bursting responses, providing a continuum of intrinsic temporal behaviors that can be harnessed at the network level.

Focusing on a challenging Mackey–Glass prediction problem pushed to its predictability limit, we showed that these richer intrinsic dynamics translate into concrete performance and efficiency gains. The MTC network substantially outperformed a standard LIF baseline in predictive fidelity, lifting the median $R^2$ from around 0.66 to approximately 0.99 and halving *SMAPE*. This outcome indicates that the multi-timescale integration of the MTC neurons provides more effective temporal memory than the single, fixed membrane time constant of the LIF, allowing the network to solve the task with fewer, more informative events.

In addition, we compared the MTC architecture with a representative SOTA AdLIF baseline. Both adaptive models achieved high accuracy, with our approach attaining slightly higher median $R^2$ and lower *SMAPE*. However, the way these models allocated spikes differed markedly. The AdLIF reached its performance through dense firing, whereas the MTC network solved the task with a ≈10× lower communication load and the highest computational sparsity among all configurations. We speculate that the heterogeneous time constants led the MTC network to encode information through brief, temporally precise bursts rather than sustained tonic activity, effectively trading a small loss in accuracy for a large gain in dynamic sparsity.

When visualized in the performance–efficiency plane of Fig. 5, the MTC network occupies a favorable upper-right region, jointly improving sparsity metrics and forecasting accuracy.

At the same time, our results also highlight important nuances and limitations. First, while AdLIF models have been shown to achieve sparser activity than LIF models on event-based classification benchmarks [16], we did not observe a substantial sparsity gap between AdLIF and LIF models in our experiments on non-spiking Mackey–Glass input. This implies that the relative advantages of adaptation versus conductance-based mechanisms may be strongly dependent on input statistics, coding scheme, and task objective. Second, our study focuses on time-series regression—an error-sensitive task relatively underrepresented in SNN benchmarks—and a feed-forward architecture; recurrent connectivity, other datasets, and more diverse input encodings may reveal different trade-offs between accuracy, sparsity, and robustness. Finally, our sparsity metrics ($S_{comp}$ and $S_{comm}$) are hardware-agnostic proxies for synaptic activity and bandwidth; while motivated by neuromorphic considerations, their quantitative relationship to true energy consumption is ultimately hardware and implementation-dependent.

Overall, the present work demonstrates that multi-timescale conductance shaping can serve as a practical and efficient building block for temporal processing in machine learning contexts. MTC neurons offer a complementary route to dynamic sparsity compared to phenomenological adaptive models: rather than tuning thresholds, adaptation strengths, or rate regularizers, sparsity in our model emerges from conductance-shaped excitability and the resulting firing regimes. These results strengthen the case for incorporating biologically grounded conductance mechanisms into SNN design, and they point to conductance-driven excitability as a strong candidate for building neuromorphic systems that are both energy-efficient and capable of rich temporal computation.

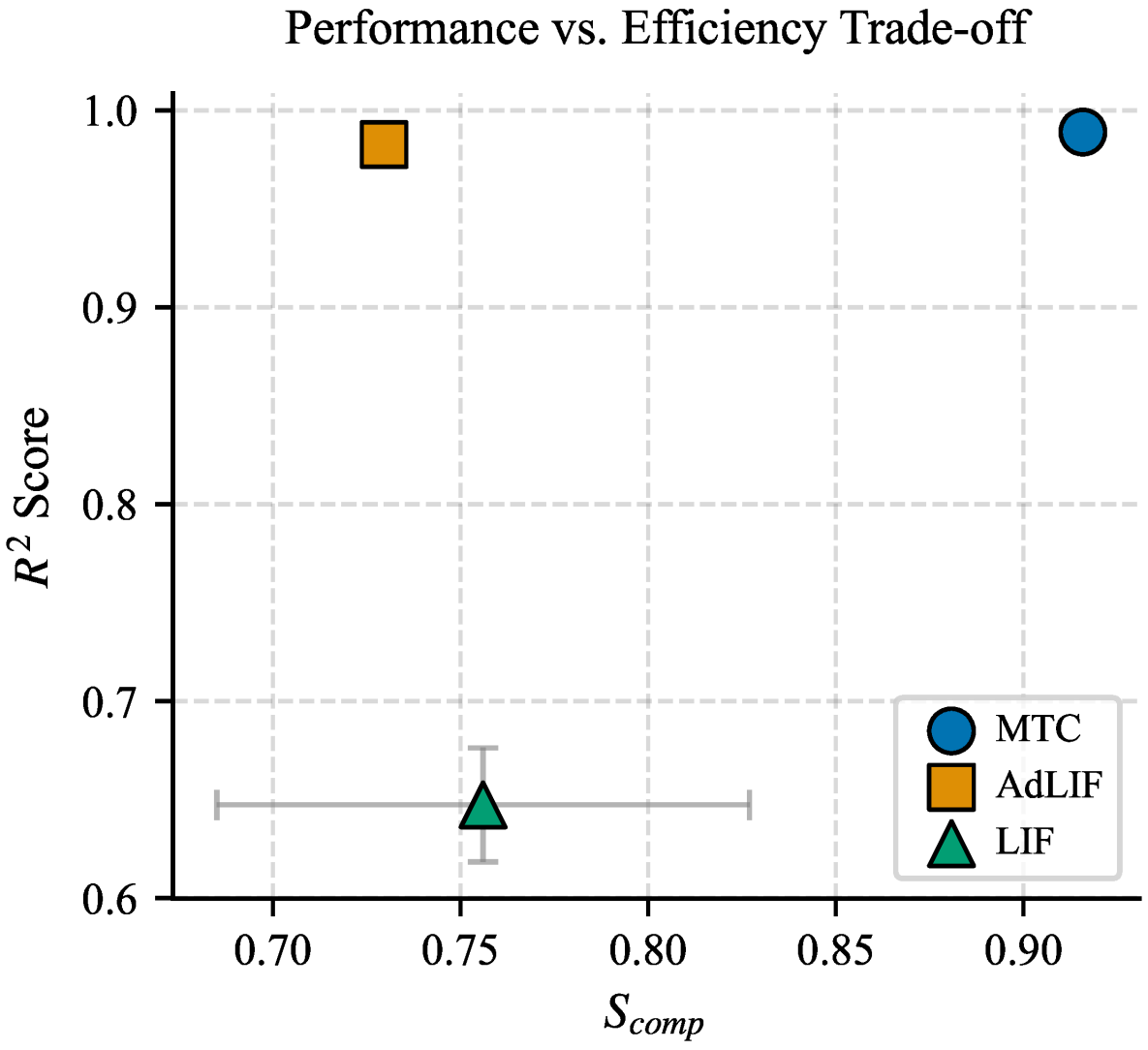

Fig. 5. Performance vs Efficiency trade-off. The proposed MTC model (blue circle) occupies the optimal upper-right quadrant, achieving high predictive fidelity ($R^2$) and synaptic computational sparsity ($S_{comp}$), in contrast to the dense firing of the AdLIF and the poor convergence of the LIF.

Several avenues remain to fully characterize and exploit the potential of MTC networks:

- **Dynamical analysis and free-run prediction.** A natural next step is to study the stability and dynamical response of MTC networks in greater depth, including free-running, autoregressive Mackey–Glass prediction at horizons beyond the one considered here. This would help disentangle open-loop prediction accuracy from closed-loop, intrinsic attractor dynamics and clarify when conductance shaping yields robust long-term behavior.
- **Spiking temporal benchmarks.** Our current evaluation uses a non-spiking time-series input. Extending the analysis to spiking temporal classification tasks such as SHD would provide a more direct comparison with the existing SNN literature and clarify how MTC neurons behave under event-based encoding and in settings where AdLIF is known to excel.
- **Broader Neuromorphic metrics and robustness.** Beyond communication and computational sparsity, MTC networks should be evaluated along additional Neurobench-style axes, including latency, memory footprint, and estimated energy consumption under realistic hardware models. Additionally, assessing robustness to noise, perturbations, and parameter variations would convey valuable information to understand the reliability of conductance-based excitability in practical deployments.
- **Initialization and architectural design;** Our current configurations use relatively simple design a mixed distribution of tonic with heterogeneous spiking neurons. A more systematic exploration of initialization strategies—including weight distributions, conductance parameter ranges, and mixtures of tonic and heterogeneous sub-populations—could yield better trainability, faster convergence, and improved performance–sparsity trade-offs.
- **Learning conductance timescales.** Finally, we have treated the time constants of the conductance elements as (largely) fixed design parameters. Allowing the network to learn these timescales jointly with the synaptic weights could unlock richer forms of temporal adaptation. Previous work has demonstrated that co-training time constants in adaptive SNNs can be beneficial for task performance [16], [18], suggesting that MTC networks could similarly benefit from gradient-based optimization of their intrinsic timescales.

Addressing these directions will be key to establishing when and how multi-timescale conductance spiking networks can deliver consistent advantages over existing SNN architectures across tasks, datasets, and neuromorphic platforms.

## References

[1] M. Davies, "Benchmarks for progress in neuromorphic computing," *Nature Machine Intelligence*, vol. 1, no. 9, pp. 386–388, 2019.

[2] J. Yik, K. Van den Berghe, D. den Blanken, Y. Bouhadjar, M. Fabre, P. Hueber, W. Ke, M. A. Khoei, D. Kleyko *et al.*, "The neurobench framework for benchmarking neuromorphic computing algorithms and systems," *Nature Communications*, vol. 16, no. 1, p. 1545, 2025.

[3] J. M. Margarit-Taulé, M. Martín-Ezquerra, R. Escudé-Pujol, C. Jiménez-Jorquera, and S.-C. Liu, “Cross-compensation of FET sensor drift and matrix effects in the industrial continuous monitoring of ion concentrations,” *Sensors and Actuators B: Chemical*, vol. 353, p. 131123, 2022.

[4] X. Li, Q. Ding, and J.-Q. Sun, “Remaining useful life estimation in prognostics using deep convolution neural networks,” *Reliability Engineering & System Safety*, vol. 172, pp. 1–11, 2018.

[5] S. Wang, M. Rovira, S. Demuru, C. Lafaye, J. Kim, B. P. Kunnel, C. Besson, C. Fernandez-Sanchez, F. Serra-Graells, J. M. Margarit-Taulé, J. Aymerich, J. Cuenca, I. Kiselev, V. Gremeaux, M. Saubade, C. Jimenez-Jorquera, D. Briand, and S.-C. Liu, “Multisensing wearables for real-time monitoring of sweat electrolyte biomarkers during exercise and analysis on their correlation with core body temperature,” *IEEE Transactions on Biomedical Circuits and Systems*, vol. 17, no. 4, pp. 808–817, 2023.

[6] T. Haarnoja, A. Zhou, P. Abbeel, and S. Levine, “Soft actor-critic: Off-policy maximum entropy deep reinforcement learning with a stochastic actor,” in *International Conference on Machine Learning*. PMLR, 2018, pp. 1861–1870.

[7] J. K. Eshraghian, M. Ward, E. Neftci, X. Wang, G. Lenz, G. Dwivedi, D. S. Bennamoun, Mohammed andhee, and W. D. Lu, “Training spiking neural networks using lessons from deep learning,” *Proceedings of the IEEE*, vol. 111, no. 9, pp. 1016–1054, 2023.

[8] Y. Wu, L. Deng, G. Li, J. Zhu, and L. Shi, “Spatio-temporal backpropagation for training high-performance spiking neural networks,” *Frontiers in Neuroscience*, vol. 12, p. 331, 2018.

[9] F. Zenke and S. Ganguli, “Superspike: Surrogate gradient learning in spiking neural networks,” *Neural Computation*, vol. 30, no. 6, pp. 1514–1541, 2018.

[10] E. M. Izhikevich, “Which model to use for cortical spiking neurons?” *IEEE Transactions on Neural Networks*, vol. 15, no. 5, pp. 1063–1070, 2004.

[11] W. Gerstner, W. M. Kistler, R. Naud, and L. Paninski, *Neuronal dynamics: From single neurons to networks and models of cognition*. Cambridge University Press, 2014.

[12] E. O. Neftci, H. Mostafa, and F. Zenke, “Surrogate gradient learning in spiking neural networks: Bringing the power of gradient-based optimization to spiking neural networks,” *IEEE Signal Processing Magazine*, vol. 36, no. 6, pp. 51–63, 2019.

[13] S. B. Shrestha and G. Orchard, “Slayer: Spike layer error reassignment in time,” in *Advances in Neural Information Processing Systems*, vol. 31, 2018.

[14] G. Bellec, D. Salaj, A. Subramoney, R. Legenstein, and W. Maass, “Long short-term memory and learning-to-learn in networks of spiking neurons,” *Advances in neural information processing systems*, vol. 31, 2018.

[15] D. Salaj, A. Subramoney, C. Kraisnikovic, G. Bellec, R. Legenstein, and W. Maass, “Spike-frequency adaptation supports network computations on temporally dispersed information,” *eLife*, vol. 10, p. e65459, 2021.

[16] B. Yin, F. Corradi, and S. M. Bohté, “Accurate and efficient time-domain classification with adaptive spiking recurrent neural networks,” *Nature Machine Intelligence*, vol. 3, no. 10, pp. 905–913, 2021.

[17] C. Ganguly, S. S. Bezugam, E. Abs, M. Payvand, S. Dey, and M. Suri, “Spike frequency adaptation: bridging neural models and neuromorphic applications,” *Communications Engineering*, vol. 3, p. 22, 2024.

[18] M. Baronig, R. Ferrand, S. Sabathiel, and R. Legenstein, “Advancing spatio-temporal processing through adaptation in spiking neural networks,” *Nature Communications*, vol. 16, no. 1, p. 5776, 2025.

[19] B. W. Connors and M. J. Gutnick, “Intrinsic firing patterns of diverse neocortical neurons,” *Trends in neurosciences*, vol. 13, no. 3, pp. 99–104, 1990.

[20] “Petilla terminology: nomenclature of features of gabaergic interneurons of the cerebral cortex,” *Nature Reviews Neuroscience*, vol. 9, no. 7, pp. 557–568, 2008.

[21] S. Soldado-Magraner, F. Brandalise, S. Honnuraiah, M. Pfeiffer, M. Moulinier, U. Gerber, and R. Douglas, “Conditioning by subthreshold synaptic input changes the intrinsic firing pattern of ca3 hippocampal neurons,” *Journal of neurophysiology*, vol. 123, no. 1, pp. 90–106, 2020.

[22] D. Debanne, Y. Inglebert, and M. Russier, “Plasticity of intrinsic neuronal excitability,” *Current opinion in neurobiology*, vol. 54, pp. 73–82, 2019.

[23] L. Ribar and R. Sepulchre, “Neuromodulation of neuromorphic circuits,” *IEEE Transactions on Circuits and Systems I: Regular Papers*, vol. 66, no. 8, pp. 3028–3040, 2019.

[24] L. Mendolia, C. Wen, E. Chicca, G. Indiveri, R. Sepulchre, J.-M. Redouté, and A. Franci, “A neuromodulable current-mode silicon neuron for robust and adaptive neuromorphic systems,” *arXiv preprint arXiv:2512.01133*, 2025.

[25] M. Mahowald and R. Douglas, “A silicon neuron,” *Nature*, vol. 354, no. 6354, pp. 515–518, 1991.

[26] G. Indiveri, B. Linares-Barranco, T. J. Hamilton, A. van Schaik, R. Etienne-Cummings, T. Delbruck, S.-C. Liu, P. Dudek, P. Häfliger, S. Renaud, J. Schemmel, G. Cauwenberghs, J. Arthur, K. Hynna, F. Folowosele, S. Saïghi, T. Serrano-Gotarredona, J. Wijekoon, Y. Wang, and K. Boahen, “Neuromorphic silicon neuron circuits,” *Frontiers in Neuroscience*, vol. 5, p. 73, 2011.

[27] M. C. Mackey and L. Glass, “Oscillation and chaos in physiological control systems,” *Science*, vol. 197, no. 4300, pp. 287–289, 1977. [Online]. Available: https://www.science.org/doi/abs/10.1126/science.267326

[28] S. Zhou, C. Gao, T. Delbruck, M. Verhelst, and S.-C. Liu, “Exploiting neuro-inspired dynamic sparsity for energy-efficient intelligent perception,” *Nature Communications*, vol. 16, no. 1, p. 9928, 2025. [Online]. Available: https://www.nature.com/articles/s41467-025-65387-7

[29] J. K. Eshraghian, “snntorch: Deep and online learning with spiking neural networks in python,” 2023, gitHub repository. [Online]. Available: https://github.com/jeshraghian/snnTorch

[30] S. Karki, D. Chavez Arana, A. Sornborger, and F. Caravelli, “Neuromorphic on-chip reservoir computing with spiking neural network architectures,” arXiv preprint arXiv:2407.20547, 2024.

[31] S. Lucas and E. Portillo, “Methodology based on spiking neural networks for univariate time-series forecasting,” *Neural Networks*, vol. 173, p. 106171, 2024.

[32] J. D. Farmer and J. J. Sidorowich, “Predicting chaotic time series,” *Phys. Rev. Lett.*, vol. 59, pp. 845–848, Aug 1987. [Online]. Available: https://link.aps.org/doi/10.1103/PhysRevLett.59.845

[33] S. Baronig, M. Schöne, A. Anand, D. Kappel, G. Parton, S. Bilgic, and R. Legenstein, “SE-adLIF: Code for "advancing spatio-temporal processing in spiking neural networks through adaptation",” 2025, gitHub repository. [Online]. Available: https://github.com/IGITUGraz/SE-adlif